\documentclass[conference]{IEEEtran}
\IEEEoverridecommandlockouts

\usepackage{amsmath,amssymb,amsfonts}
\usepackage{algorithmic}
\usepackage{graphicx}
\usepackage{textcomp}
\usepackage{xcolor}
\def\BibTeX{{\rm B\kern-.05em{\sc i\kern-.025em b}\kern-.08em
    T\kern-.1667em\lower.7ex\hbox{E}\kern-.125emX}}

\usepackage{soul}
\usepackage{url}
\usepackage[hidelinks]{hyperref}
\usepackage[utf8]{inputenc}
\usepackage[small]{caption}
\usepackage{amsthm}
\usepackage{booktabs}
\usepackage{algorithm}
\usepackage[switch]{lineno}
\usepackage{pgfplots}
\pgfplotsset{compat = newest}
\usepgfplotslibrary{groupplots}
\usepgfplotslibrary{colorbrewer}
\usepackage{tikz}
\usetikzlibrary{matrix, positioning,external}
\usepackage{circuitikz}
\usepackage{standalone}
\usepackage{xcolor}
\usepackage{cite}
\usepackage{tabularray}
\usepackage{multirow}
\usepackage{booktabs}
\usepackage{subcaption}
\UseTblrLibrary{booktabs} %
\usepackage{float}

\usepackage{xcolor}

\definecolor{darkWhite}{rgb}{0.96,0.96,0.96}
\definecolor{bluekeywords}{rgb}{0.13,0.13,1}
\definecolor{greencomments}{rgb}{0,0.5,0}
\definecolor{redstrings}{rgb}{0.9,0,0}

\definecolor{Comment}{RGB}{97,161,176}

\definecolor{btfGreen}{RGB}{51,160,44}
\definecolor{btfRed}{RGB}{190,60,90}

\definecolor{bleuUni}{RGB}{0, 157, 224}
\definecolor{marronUni}{RGB}{68, 58, 49}

\definecolor{bluecite}{HTML}{009DE0}

\definecolor{Paired_1}{RGB}{31,120,180} 
\definecolor{Paired_2}{RGB}{166,206,227} 
\definecolor{Paired_3}{RGB}{51,160,44} 
\definecolor{Paired_4}{RGB}{178,223,138} 
\definecolor{Paired_5}{RGB}{227,26,28} 
\definecolor{Paired_6}{RGB}{251,154,153} 
\definecolor{Paired_7}{RGB}{255,127,0} 
\definecolor{Paired_8}{RGB}{253,191,111} 
\definecolor{Paired_9}{RGB}{106,61,154} 
\definecolor{Paired_10}{RGB}{202,178,214} 
\definecolor{Paired_11}{RGB}{177,89,40} 
\definecolor{Paired_12}{RGB}{255,255,153} 
\definecolor{Accent_1}{RGB}{127,201,127} 
\definecolor{Accent_2}{RGB}{190,174,212} 
\definecolor{Accent_3}{RGB}{253,192,134} 
\definecolor{Accent_4}{RGB}{255,255,153} 
\definecolor{Accent_5}{RGB}{56,108,176} 
\definecolor{Accent_6}{RGB}{240,2,127} 
\definecolor{Accent_7}{RGB}{191,91,23} 
\definecolor{Accent_8}{RGB}{102,102,102} 
\definecolor{Spectral_1}{RGB}{158,1,66} 
\definecolor{Spectral_2}{RGB}{213,62,79} 
\definecolor{Spectral_3}{RGB}{244,109,67} 
\definecolor{Spectral_4}{RGB}{253,174,97} 
\definecolor{Spectral_5}{RGB}{254,224,139} 
\definecolor{Spectral_6}{RGB}{255,255,191} 
\definecolor{Spectral_7}{RGB}{230,245,152} 
\definecolor{Spectral_8}{RGB}{171,221,164} 
\definecolor{Spectral_9}{RGB}{102,194,165} 
\definecolor{Spectral_10}{RGB}{50,136,189} 
\definecolor{Spectral_11}{RGB}{94,79,162} 
\definecolor{Set1_1}{RGB}{228,26,28} 
\definecolor{Set1_2}{RGB}{55,126,184} 
\definecolor{Set1_3}{RGB}{77,175,74} 
\definecolor{Set1_4}{RGB}{152,78,163} 
\definecolor{Set1_5}{RGB}{255,127,0} 
\definecolor{Set1_6}{RGB}{255,255,51} 
\definecolor{Set1_7}{RGB}{166,86,40} 
\definecolor{Set1_8}{RGB}{247,129,191} 
\definecolor{Set1_9}{RGB}{153,153,153} 
\definecolor{Set1_10}{RGB}{0,0,0} 
\definecolor{Set2_1}{RGB}{102,194,165} 
\definecolor{Set2_2}{RGB}{252,141,98} 
\definecolor{Set2_3}{RGB}{141,160,203} 
\definecolor{Set2_4}{RGB}{231,138,195} 
\definecolor{Set2_5}{RGB}{166,216,84} 
\definecolor{Set2_6}{RGB}{255,217,47} 
\definecolor{Set2_7}{RGB}{229,196,148} 
\definecolor{Set2_8}{RGB}{179,179,179} 
\definecolor{Dark2_1}{RGB}{27,158,119} 
\definecolor{Dark2_2}{RGB}{217,95,2} 
\definecolor{Dark2_3}{RGB}{117,112,179} 
\definecolor{Dark2_4}{RGB}{231,41,138} 
\definecolor{Dark2_5}{RGB}{102,166,30} 
\definecolor{Dark2_6}{RGB}{230,171,2} 
\definecolor{Dark2_7}{RGB}{166,118,29} 
\definecolor{Dark2_8}{RGB}{102,102,102} 
\definecolor{Reds_1}{RGB}{255,245,240} 
\definecolor{Reds_2}{RGB}{254,224,210} 
\definecolor{Reds_3}{RGB}{252,187,161} 
\definecolor{Reds_4}{RGB}{252,146,114} 
\definecolor{Reds_5}{RGB}{251,106,74} 
\definecolor{Reds_6}{RGB}{239,59,44} 
\definecolor{Reds_7}{RGB}{203,24,29} 
\definecolor{Reds_8}{RGB}{165,15,21} 
\definecolor{Reds_9}{RGB}{103,0,13} 
\definecolor{Greens_1}{RGB}{247,252,245} 
\definecolor{Greens_2}{RGB}{229,245,224} 
\definecolor{Greens_3}{RGB}{199,233,192} 
\definecolor{Greens_4}{RGB}{161,217,155} 
\definecolor{Greens_5}{RGB}{116,196,118} 
\definecolor{Greens_6}{RGB}{65,171,93} 
\definecolor{Greens_7}{RGB}{35,139,69} 
\definecolor{Greens_8}{RGB}{0,109,44} 
\definecolor{Greens_9}{RGB}{0,68,27} 
\definecolor{Blues_1}{RGB}{247,251,255} 
\definecolor{Blues_2}{RGB}{222,235,247} 
\definecolor{Blues_3}{RGB}{198,219,239} 
\definecolor{Blues_4}{RGB}{158,202,225} 
\definecolor{Blues_5}{RGB}{107,174,214} 
\definecolor{Blues_6}{RGB}{66,146,198} 
\definecolor{Blues_7}{RGB}{33,113,181} 
\definecolor{Blues_8}{RGB}{8,81,156} 
\definecolor{Blues_9}{RGB}{8,48,107} 

\title{Input Resolution Downsizing as a Compression Technique for Vision Deep Learning Systems}

\author{\IEEEauthorblockN{1\textsuperscript{st} Jérémy Morlier}
\IEEEauthorblockA{\textit{IMT Atlantique} \\
\textit{Lab-STICC, UMR CNRS 6285}\\
F-29238 Brest, France \\
jeremy.morlier@imt-atlantique.fr}
\and
\IEEEauthorblockN{2\textsuperscript{nd} Mathieu Léonardon}
\IEEEauthorblockA{\textit{IMT Atlantique} \\
\textit{Lab-STICC, UMR CNRS 6285}\\
F-29238 Brest, France \\
mathieu.leonardon@imt-atlantique.fr}
\and
\IEEEauthorblockN{3\textsuperscript{rd} Vincent Gripon}
\IEEEauthorblockA{\textit{IMT Atlantique} \\
\textit{Lab-STICC, UMR CNRS 6285}\\
F-29238 Brest, France \\
vincent.gripon@imt-atlantique.fr}
}
\begin{document}

\maketitle

\begin{abstract} 

Model compression is a critical area of research in deep learning, in particular in vision, driven by the need to lighten models memory or computational footprints. While numerous methods for model compression have been proposed, most focus on pruning, quantization, or knowledge distillation. In this work, we delve into an under-explored avenue: reducing the resolution of the input image as a complementary approach to other types of compression. By systematically investigating the impact of input resolution reduction, on both tasks of classification and semantic segmentation, and on convnets and transformer-based architectures, we demonstrate that this strategy provides an interesting alternative for model compression. Our experimental results on standard benchmarks highlight the potential of this method, achieving competitive performance while significantly reducing computational and memory requirements. This study establishes input resolution reduction as a viable and promising direction in the broader landscape of model compression techniques for vision applications.
\end{abstract}


\section{Introduction} 

Reducing the size of deep learning models has become a critical area of research, particularly in computer vision, where large models often require significant memory and computational resources. As deep learning continues to find applications in resource-constrained environments, such as mobile devices and embedded systems, efficient model compression techniques are essential for maintaining model usability without compromising performance. Reducing the size of models is also beneficial in the context of data centers.

Over the years, a variety of model compression methods have been proposed in the literature, including pruning~\cite{han2015learning, frankle2018lottery
}, quantization~\cite{Jacob_2018_CVPR, hubara2016binarized}, knowledge distillation~\cite{hinton2015distilling}, and low-rank approximation~\cite{denton2014exploiting}.

\begin{figure}
    \centering
    \resizebox{0.4\textwidth}{!}{
    {
\begin{tikzpicture}

\definecolor{darkgray176}{RGB}{176,176,176}
\definecolor{darkorange25512714}{RGB}{255,127,14}
\definecolor{forestgreen4416044}{RGB}{44,160,44}
\definecolor{lightgray204}{RGB}{204,204,204}
\definecolor{steelblue31119180}{RGB}{31,119,180}

\begin{axis}[
legend cell align={left},
legend style={
  fill opacity=0.8,
  draw opacity=1,
  text opacity=1,
  at={(0.99,0.01)},
  anchor=south east,
  draw=lightgray204
},
xmode=log,
ymode=log,
xminorgrids=true,
yminorgrids=true,
tick align=outside,
tick pos=left,
xlabel={Required Memory (MBs)},
ylabel={Computational Cost (GFLOPs)},
xmajorgrids,
x grid style={darkgray176},
xtick style={color=black},
xmin=9e6, xmax=1.2e8,
xtick={1e7, 1e8},
xticklabels={10, 100},
scaled x ticks=false,
ymajorgrids,
y grid style={darkgray176},
ytick style={color=black},
ymin=3e9, ymax=10e10,
ytick={10e8, 10e9, 100e9},
yticklabels={1, 10, 100},
scaled y ticks=false,
]
\addplot [draw=steelblue31119180, fill=steelblue31119180, mark size=2, mark=*, only marks] table [x=a, y=b,col sep=comma]  {figures/Resnet0/points3.csv};\label{pgfplots:resnet0_0}
\addplot [draw=forestgreen4416044, fill=forestgreen4416044, mark size=2, mark=*, only marks] table [x=e, y=f,col sep=comma]  {figures/Resnet0/points3.csv};\label{pgfplots:resnet0_1}
\addplot [draw=darkorange25512714, fill=darkorange25512714, mark size=2, mark=*, only marks] table [x=c, y=d,col sep=comma]  {figures/Resnet0/points3.csv};\label{pgfplots:resnet0_2}
\addplot [semithick, black, dashed] table [x=g, y=h,col sep=comma] {figures/Resnet0/resnet3.csv};\label{pgfplots:resnet0_3}

\addplot [draw=steelblue31119180, fill=steelblue31119180, mark size=2, mark=square*, only marks] table [x=g, y=h,col sep=comma]  {figures/Resnet0/points3.csv};\label{pgfplots:resnet0_4}
\addplot [draw=forestgreen4416044, fill=forestgreen4416044, mark size=2, mark=square*, only marks] table [x=k, y=l,col sep=comma]  {figures/Resnet0/points3.csv};\label{pgfplots:resnet0_5}
\addplot [draw=darkorange25512714, fill=darkorange25512714, mark size=2, mark=square*, only marks] table [x=i, y=j,col sep=comma]  {figures/Resnet0/points3.csv};\label{pgfplots:resnet0_6}
\coordinate (legend) at (axis description cs:1.0,0.0);
\end{axis}
\matrix [
    draw,
    matrix of nodes,
    anchor=south east,
    fill=white,
    row sep=-2pt,
    column sep=-9pt,
    style={nodes={font=\small}}
] at (legend) {
    \ref{pgfplots:resnet0_3} \hspace{1pt} \pgfmatrixnextcell  Model Scaling   \\
     MS \pgfmatrixnextcell Accuracy        \pgfmatrixnextcell RS             \\ 
    \ref{pgfplots:resnet0_0} \pgfmatrixnextcell $\simeq79\%$ \pgfmatrixnextcell \ref{pgfplots:resnet0_4}\\
    \ref{pgfplots:resnet0_1} \pgfmatrixnextcell $\simeq74\%$ \pgfmatrixnextcell \ref{pgfplots:resnet0_5}\\
    \ref{pgfplots:resnet0_2} \pgfmatrixnextcell $\simeq65\%$ \pgfmatrixnextcell \ref{pgfplots:resnet0_6} \\
};

\end{tikzpicture}

    }}
    \caption{Benefits of adding resolution scaling to the more classical model scaling, when performing classification on ImageNet using a ResNet-50 with a batch size of 8. The dotted line represents the achievable trade-offs when using model scaling only. Three circle points correspond to various model scaling (MS), achieving different accuracy levels. The square points of the same colors correspond to adding input resolution scaling (RS), achieving a better trade-off without sacrificing accuracy.
}
    \label{fig:catchFigure}
\end{figure}
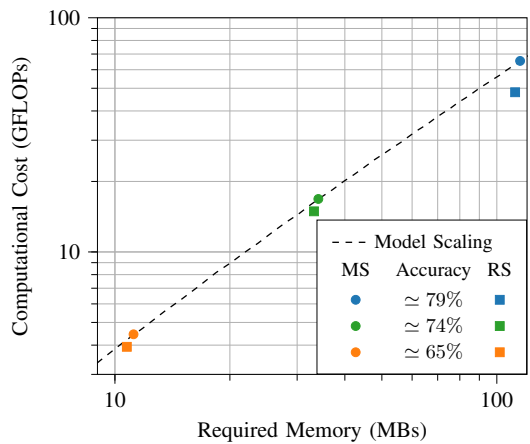 

In this context, changing the resolution of the input images is a relatively less used model compression technique that can be used post-training. Intuitively, reducing the resolution of input images can significantly lower computational cost and memory needs, because it directly impacts the size of feature maps in convolutional-based architectures. It is also a simple way to reduce the size of the sequence of tokens in transformer-based architectures. Interestingly, the trade-offs between performance degradation and computational savings in this approach have not been thoroughly investigated, especially across diverse tasks and architectures.

In this work, we delve into the potential of input resolution reduction as an additional strategy for model compression to moderately reduce the computational cost and memory needs. We investigate this method across two fundamental problems in computer vision: image classification and semantic segmentation, and consider two prominent types of architectures: convolutional neural networks (e.g., ResNets~\cite{he2016deep}) and transformer-based architectures (e.g., Vision Transformers~\cite{dosovitskiyImageWorth16x162021}).

Our problem statement centers on whether input resolution reduction can serve as a viable alternative to model scaling strategies. Specifically, we aim to evaluate its effectiveness in striking a balance between computational or memory efficiency and task performance, offering insights into whether it complements or even outperforms conventional model compression methods. We make several contributions in this study: 

\begin{itemize}\item We propose mechanisms for modifying input resolution both before and after the embedding process in ResNets and for reducing the length of the sequence of tokens in Vision Transformers, \item We systematically examine the potential of input resolution reduction as a standalone or complementary compression technique with model scaling or quantization, \item We present experiments demonstrating that resolution reduction consistently provides a better trade-off between performance and computational cost than established alternatives on multiple problems and architectures.\end{itemize}
Our findings underscore the importance of considering input resolution as a practical dimension in the broader landscape of model compression for computer vision.

\section{Related Work} 
Several works have explored the use of resolution as a scaling factor in deep learning architectures, often distinguishing their approaches based on their application to CNNs or ViTs.

For CNNs, image resolution is a critical factor influencing the design and performance of deep learning models across various tasks with sizes ranging from 224x224 for image classification to 1024x1024 for semantic segmentation. Resolution as a scaling factor for compression was introduced by the MobileNet architectures~\cite{howard2017mobilenets} as a new method for reducing the computational cost of deep learning architectures.
EfficientNet~\cite{tanEfficientNetRethinkingModel2020} proposed a neural architecture search method combining width, depth and resolution on an efficient MobileNetv2 based~\cite{sandler2018mobilenetv2} architecture. Similarly,~\cite{dollar2021fast} introduces compound scaling laws on EfficientNet and RegNet architectures in order to easily scale up these architectures. In~\cite{SizeMatters, touvronFixingTraintestResolution2022}, the authors highlight the different impacts the resolution of the input image can have on the design, the training and the evaluation of a convolutional deep learning system.

For ViTs, exploiting patch size has been applied as a scale up technique as finer patch size allows to improve performance on fine-grained tasks. Swin transformers~\cite{liu2021swin}, PiT~\cite{heo2021rethinking} employ smaller patch size of 8x8 pixels in order to catch finer-grained details at the beggining of the network and pool them at a later stage in the architecture to reduce the increased computational cost of using smaller patch size. 
VITAR~\cite{fan2024vitar} and UniVit~\cite{likhomanenko2021cape} both focus on enhancing positional embeddings to improve transformer performance at varying input resolutions. VITAR introduces a fuzzy positional encoding combined with a token-merging mechanism to adapt input sequences for neural network processing. UniVit employs augmented positional embeddings to enhance generalization across resolutions and resizes image batches to random dimensions within the range $\left[128, 320\right]$ with 32-pixel increments during training. This resolution generalization strategy has also been applied in DinoV2~\cite{oquab2023dinov2}, an image foundation model.
Our work shows that resolution should not only be viewed as a training-time augmentation factor or as part of a compound scaling rule, but also as a practical and effective standalone compression dimension for both CNNs and ViTs.

\section{Methodology} 

\subsection{Motivation} 
As mentioned in the introduction, the question of compressing -- meant as reducing the number of required computations and memory -- a deep learning architecture has known a large number of contributions in the past decade~\cite{hinton2015distilling, nagel2021white, jacob2017antization, he2017channel, han2015deep}.

Popular compression techniques such as pruning, quantization, distillation or model scaling are widely studied in order to compress architectures for inference. 
Introduced in~\cite{tanEfficientNetRethinkingModel2020} for CNNs, model scaling was initially thought of as a combination of width, depth and input resolution. Outside the context of EfficientNets, it is common to act upon both width and depth when designing an architecture for a given problem~\cite{zhai2022scaling,chen2021searching}.
Interestingly, we observe that the community disregards acting upon the input resolution of considered images for compression purposes~\cite{gordon2018morphnet, lee2020neuralscale}, especially for ViTs~\cite{zhai2022scaling, tang2024survey}. In this paper, we aim to systematically explore the trade-offs between performance and complexity across various tasks and architecture families, namely classification and semantic segmentation for CNNs and ViTs. As explained later, we can draw an analogy between resolution scaling for CNNs and sequence length for ViTs.

We demonstrate, on one hand, that the trade-offs between computational complexity and memory usage achievable through resolution scaling differ significantly from those achievable through model scaling. On the other hand, we show that in many cases, the performance-to-complexity trade-off favors resolution scaling over model scaling.

From the outset, it should be noted that this study excludes comparisons with pruning techniques. Indeed, only structured pruning directly reduces the computational complexity of neural networks. Structured pruning refers to the removal of entire filters in convolutional layers or complete rows/columns in the matrix computations of transformers. Under this constraint, we observe that comparisons between model scaling and pruning generally disfavor the latter~\cite{tessier2023thinresnet, liu2018rethinking, crowley2018closer}.

Additionally, we consider quantization techniques on convolutional neural networks that shows that quantization can be combined with both model scaling and resolution scaling. Supporting results are provided in~\ref{quantif}.

\begin{figure}[t]
    \centering
    \resizebox{0.4\textwidth}{!}{
    {
\begin{tikzpicture}

\definecolor{darkgray176}{RGB}{176,176,176}
\definecolor{lightgray204}{RGB}{204,204,204}
\begin{axis}[
legend cell align={left},
legend style={
  fill opacity=0.8,
  draw opacity=1,
  text opacity=1,
  at={(0.03,0.97)},
  anchor=north west,
  draw=lightgray204
},
tick align=outside,
tick pos=left,
xlabel={Memories Required (MBs)},
ylabel={Computational Cost (GFLOPs)},
xmajorgrids,
x grid style={darkgray176},
xtick style={color=black},
xmin=0, xmax=120000000,
xtick={0.0, 20000000, 40000000, 60000000, 80000000, 100000000, 120000000, 140000000},
xticklabels={0.0, 20, 40, 60, 80, 100, 120, 140},
scaled x ticks=false,
ymajorgrids,
y grid style={darkgray176},
ytick style={color=black},
ymin=0, ymax=10000000000,
ytick={0.0, 2000000000, 4000000000, 6000000000, 8000000000, 10000000000},
yticklabels={0.0, 2, 4, 6, 8, 1},
scaled y ticks=false,
]

\addplot [thick, Set1_3, dashed, mark=*, mark size=1.5, mark options={solid}] table [x=MSM, y=MSF,col sep=comma] {figures/SearchSpaceCNN/SearchSpace.csv};
\addlegendentry{Model Scaling}
\addplot [thick, Set1_2, dashed, mark=*, mark size=1.5, mark options={solid}] table [x=RSM, y=RSF,col sep=comma] {figures/SearchSpaceCNN/SearchSpace.csv};
\addlegendentry{Resolution Scaling}
\end{axis}
\end{tikzpicture}
}
}
    \vspace{-25pt}
    \caption{Comparison of memory and FLOPs trade-offs for model scaling versus resolution scaling in ResNet-50.}
    \label{fig:SearchSpace1}
\end{figure}
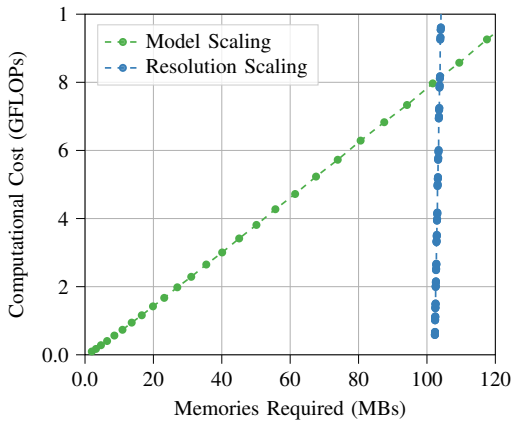 

\begin{figure*}[hbtp]
    \centering
    \includegraphics[width=\linewidth]{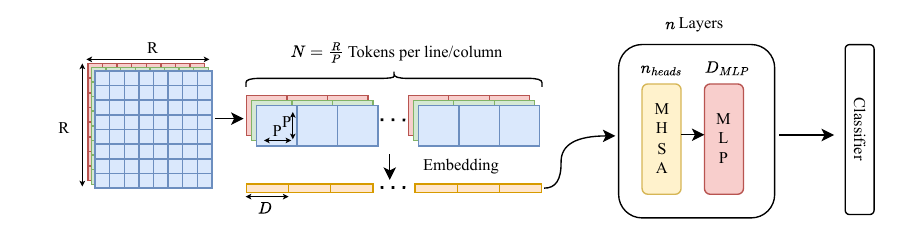}
    \label{vitSchema}
    \caption{Overview of the Vision transformer (ViT) architecture, The input image, with dimensions $R \times R$, is divided into non-overlapping patches of size $P \times P$, resulting in an input resolution of $N=\frac{R}{P}$ tokens per line/column and a total sequence length of $N^{2}$. Each token is flattened and projected into a $D$-dimensional embedding. These embeddings are then fed into the transformer model, consisting of $n$ layers, where each layer includes a Multi-Head Self-Attention (MHSA) module with $n_{heads}$ heads and a Multi-Layer Perceptron (MLP) with dimensionality $D_{MLP}$. The output is passed to a classifier for the final prediction.}
    
\end{figure*}

\subsection{Costs in computations and memory}

There is no consensual way to measure computations and memory requirements for estimating the computational complexity of a deep learning architecture~\cite{ivanov2021data, thompson2020computational, wang2019benchmarking}.
Indeed, the purpose of compression, which generally involves physical metrics related to throughput, response time, or energy consumption, may depend on various factors that are more or less complex and intertwined.
In this paper, we decide to focus on two simplified metrics, one for computations and one for memory, that we describe thereafter.

For computations, we chose to estimate the number of FLOPs (floating-point operations), a commonly adopted metric in the compression literature~\cite{tanEfficientNetV2SmallerModels2021}. FLOPs, despite disregarding memory accesses~\cite{ivanov2021data} and ignoring the difference in cost between various operators, are loosely related to computation time and energy consumption.

\begin{figure*}[htp]
    \centering
    \includegraphics{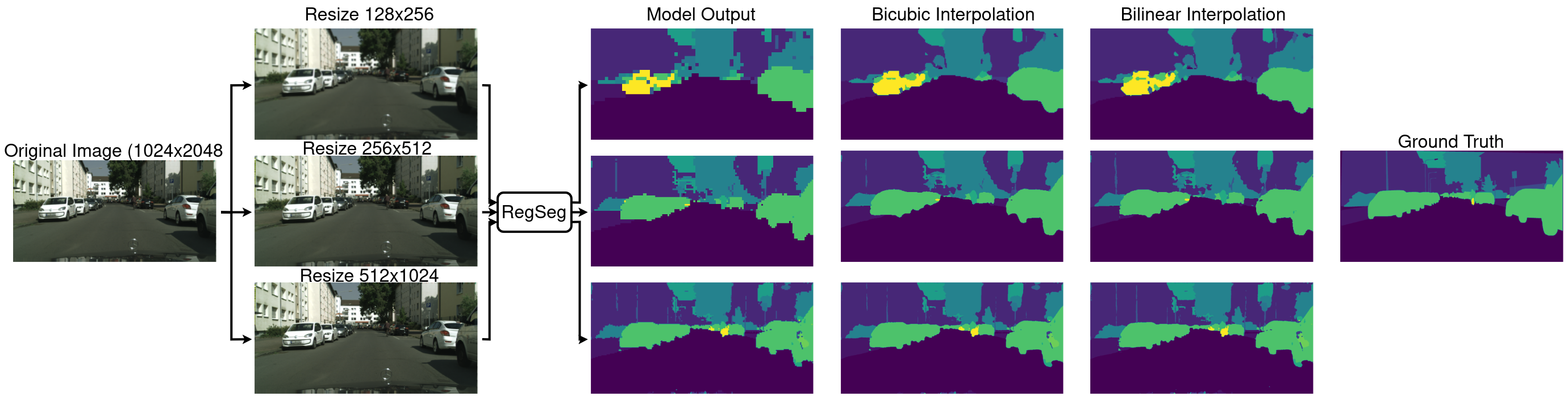}
    \caption{Illustration of the impact of resolution scaling on a RegSeg model. On the left column, several resize resolutions (128x256 to 512x1024) are applied to the original image followed by their respective model outputs. An interpolation technique such as bicubic or bilinear is then applied on the model output and the mIoU is calculated based on the ground truth.}
    \label{SchemaRegSeg}
\end{figure*}

\begin{figure}[hb]
    \centering
    \resizebox{0.4\textwidth}{!}{{
    \begin{tikzpicture}
        \definecolor{darkgray176}{RGB}{176,176,176}
        \definecolor{goldenrod1911910}{RGB}{191,191,0}
        \definecolor{green01270}{RGB}{0,127,0}
        \definecolor{lightgray204}{RGB}{204,204,204}
        
        \begin{axis}[
        legend cell align={left},
        legend style={
          draw=black,
          draw opacity=1,
          at={(0.0,1)},
          anchor=north west,
          draw=lightgray204
        },
        tick align=outside,
        tick pos=left,
        x grid style={darkgray176},
        xlabel={Memory Required (MBs)},
        xmajorgrids,
        xmin=0, xmax=90000000,
        xtick style={color=black},
        xtick={0,10000000,20000000,30000000,40000000,50000000,60000000,70000000,80000000,90000000},
        xticklabels={0,10,20,30,40,50,60,70,80,90},
        y grid style={darkgray176},
        ylabel={Computational Cost (GFLOPs)},
        ymajorgrids,
        ymin=0, ymax=11000000000,
        ytick style={color=black},
        ytick={0,2000000000,4000000000,6000000000,8000000000,10000000000,12000000000},
        yticklabels={0.0,2,4,6,8,10,12},
        scaled x ticks=false,
        scaled y ticks=false,
        ]
        \addplot [semithick, red, dashed, mark=*, mark size=1.5, mark options={solid}] table [x=a, y=b,col sep=comma]   {figures/SearchSpaceViT/SearchSpace2.csv};\label{plot:vit11}
        \addlegendentry{Sequence scaling}
        \addplot [semithick, blue, dashed, mark=*, mark size=1.5, mark options={solid}] table [x=c, y=d,col sep=comma]   {figures/SearchSpaceViT/SearchSpace2.csv};\label{plot:vit11}
        \addlegendentry{Dim scaling}
        \addplot [semithick, green01270, dashed, mark=*, mark size=1.5, mark options={solid}] table [x=e, y=f,col sep=comma]   {figures/SearchSpaceViT/SearchSpace2.csv};\label{plot:vit11}
        \addlegendentry{MLP scaling}
        \addplot [semithick, goldenrod1911910, dashed, mark=*, mark size=1.5, mark options={solid}] table [x=g, y=h,col sep=comma]   {figures/SearchSpaceViT/SearchSpace2.csv};\label{plot:vit11}
        \addlegendentry{Layers scaling}
        \end{axis}
    \end{tikzpicture}
}}
    \vspace{-25pt}
    \caption{Comparison of the memory required and the number of FLOPs trade-offs between model scaling and sequence scaling for a ViT-S.}
    \label{fig:SearchSpaceViT1}
\end{figure}
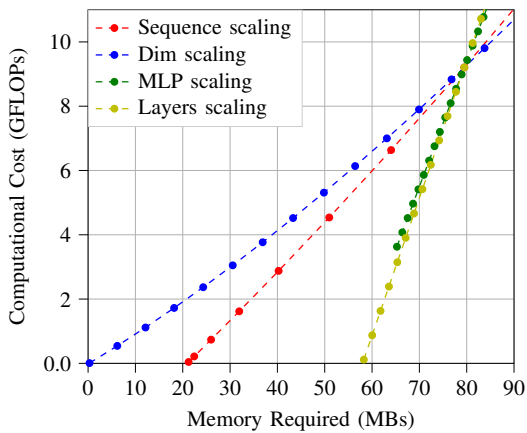

For memory, we consider the sum of the model size and the memory required to store the activations. We choose to estimate the latter using the largest memory needed to store any operators of the considered architecture as the sum of its inputs and output sizes.
This can be used as a lower theoretical lower bound~\cite{pisarchyk2020efficient} for the memory required to store the activations during an inference, even though it may not be feasible in practice.

\subsection{Scaling CNNs and ViTs}

Modern architectures for vision include both CNNs and ViTs. Since their processing are distinct, we elaborate on the impact of input resolution for each of them. Note that these architectures can contain a linear layer as their final layer, meant to project representations to a decision on the considered task, but that this layer typically bears a negligible impact on both computations and memory overall. Consequently, the computations and memory requirements of CNNs and ViTs are primarily determined by their convolutional layers and ViT blocks, respectively. ViT blocks are typically composed of multi headed attention layers and MLP layers. 
Note that the input resolution refers to distinct concepts in CNNs and ViTs. In CNNs, raw images are processed, and as such the input resolution refers to the number of pixels on a given line/column of this image. In ViTs, tokens are obtained from the initial image and then processed throughout the architecture. As such, the input resolution of ViTs refers to the number of tokens per line/column of the input image.

\textbf{CNNs:} In CNNs, the number of FLOPs in a convolutional layer (disregarding strides) is proportional to the number of incoming channels, outcoming channels, and the square of the kernel size and incoming resolution. Memory activations are proportional to the square of the incoming resolution and the sum of incoming and outcoming channels. The size of the model remains unaffected by the input resolution and is proportional to the product of the input and output channel dimensions.
As such, acting on the input resolution has a quadratic impact on both computations and memory with an offset, whereas acting on the width (here the number of channels) has a quadratic impact on computations but only linear on memory. This difference in the attainable search space between using width-based model scaling vs. resolution scaling is depicted in Figure~\ref{fig:SearchSpace1} when considering a ResNet-50.

\textbf{ViTs:} In ViTs, let us denote $k$ the number of heads, $D$ the inner dimension, $D_{MLP}$ the hidden dimension in MLPs, and $N$ the number of tokens per line/column in input images, which correspond to our input resolution. Disregarding the embedding layer, the number of FLOPs of a ViT block is: $4 N^4 D + 3 k N^4 + 2 N^2 D^2 + 4 N^2 D D_{MLP}$~\cite{kaplan2020scaling, hoffmann2022training}. For the activations, we consider that some of the multi-head self-attention operations are fused using Flash attention~\cite{dao2022flashattention} meaning that the only tensors that need to be kept in memory are the key, queries and values tensors as well as the output sequence resulting in the following cost for the activations memory: $5 N^2 D + N^2 D_{MLP}$. For the model size, we do not take into account the embedding and the classifier layer for simplicity as they do not result in an important amount of the model size, resulting in a ViT block memory footprint of $D.(3.D + D + 2.D_{MLP})$.
We observe that the reducing the input resolution  has a complexity of $\mathcal{O}(N^{4})$ on the computational cost compared to memory with a complexity of $\mathcal{O}(N^{2})$. No other hyperparameters exhibit the same relation between FLOPs and memory. This difference in the attainable search space between using width-based model scaling (here $k$, $D$ or $D_{MLP}$) and resolutions scaling is depicted in Figure~\ref{fig:SearchSpaceViT1} when considering a ViT-S architecture.

For both CNNs and ViTs, batch size is another crucial factor that scales linearly with memory and computation, while leaving model size unchanged. Increasing batch size amplifies the relative differences between model scaling and resolution scaling, making resolution scaling increasingly favorable as batch size grows.

\subsection{Methods}\label{methodo:methods}

As CNNs and ViTs use different processings for images,  we consider different methods to apply resolution scaling to those architectures.

\textbf{CNNs}: CNNs typically employs a image preprocessing pipeline that is different in training and evaluation. The training pipeline is mainly composed of a random resize crop of size $K_{train}$ and the evaluation pipeline is mainly composed of a random resize of size $K$ followed by a central crop of size $K_{eval}$. As shown in~\cite{touvronFixingTraintestResolution2022}, this difference of pipeline implies a discrepancy for the neural network between the training phase and the evaluation phase resulting in the best resolution $K_{eval}$ in evaluation being higher that the one used during training $K_{train}$. 
Using this discrepancy, we propose to reduce the random crop size used during training in order to train a CNN that can be more effective at a lower evaluation crop size and thus reducing its cost in compute and in memory.  

For our second method, we propose to amplify this discrepancy effect by employing a higher training resolution, then reducing the resolution through downsampling the activations at deeper layers.

\textbf{ViTs}: ViTs typically employs a similar preprocessing pipeline as CNNs but with the the evaluation crop size $K$ being equal to the training crop size $K_{train}$. As previously explained, we train ViTs for specific resolution using a fixed number of tokens, effectively controlling computational and memory costs. As the images are divided in patches in order to create the sequence of tokens, the size of patches is another important factor in the design of a ViT for a specific resolution (i.e. number of tokens). In this paper, we also evaluate the relationship between the patch size and the sequence length.

The proposed methodology consists in systematically evaluating the  relationship between the input resolution and the patch size in order  to lessen computations and memory.

\section{Experiments} 

\subsection{Convolutional neural networks}

In this section, we evaluate our methods on convolutional neural networks on two tasks, classification and semantic segmentation.

\subsubsection{Classification}\label{resnet}

For classification tasks, we select the ResNet-50 model on the ImageNet dataset as it is a widely studied task.
We train every ResNet-50 with the state-of-the-art training routine from torchvision~\cite{Pytorch}, that mainly uses a data augmentation transformation composed of a random crop of size $K_{train}$ with $176$ as the baseline. The standard evaluation procedure on ImageNet consists of a sequence of transformations: normalization, resizing to size $K$, and a central crop of size $K_{eval}$ with baseline values of respectively $232$ and $224$.

We evaluate our proposed methods using a setup that allows a fair comparison between the two methods. 
For the first method, we train ResNet-50 models with random crop sizes $K_{train}$ in \{$64$, $128$, $160$, $176$\}, and evaluate each network using various $K$ and $K_{eval}$ values. To compare the two methods, the second approach trains the model using a random crop size of $224$, with the activations after the first convolution resized to match the lower resolutions corresponding to the first method. This approach ensures similar training costs across the two methods, as most activation shapes remain unchanged during training.

\begin{figure}[hbtp]
    \centering
    \resizebox{0.4\textwidth}{!}{
    {
\begin{tikzpicture}

\definecolor{darkgray176}{RGB}{176,176,176}
\definecolor{lightgray204}{RGB}{204,204,204}
\definecolor{steelblue31119180}{RGB}{31,119,180}

\begin{axis}[
legend cell align={right},
legend style={
  fill opacity=0.8,
  draw opacity=1,
  text opacity=1,
  at={(0.0,1)},
  anchor=north west,
  draw=lightgray204
},
tick align=outside,
tick pos=left,
x grid style={darkgray176},
xlabel={Computational Complexity as $K_{eval}$},
xmajorgrids,
xmin=100, xmax=365.5,
xtick style={color=black},
y grid style={darkgray176},
ylabel={ImageNet Top-1 accuracy (\%)},
ymajorgrids,
ymin=76, ymax=82,
ytick style={color=black}
]
\addplot [thick, Greens_6, mark=*, mark size=1.5, mark options={solid}] table [x=a, y=c,col sep=comma] {figures/Resnet1/resnet1.csv};
\addlegendentry{Baseline}
\addplot [thick, Greens_6, mark=x, mark size=1.5, mark options={solid}] table [x=a, y=d,col sep=comma] {figures/Resnet1/resnet1.csv};
\addlegendentry{FCR $88$}
\addplot [thick, Blues_6, mark=*, mark size=1.5, mark options={solid}] table [x=a, y=f,col sep=comma] {figures/Resnet1/resnet1.csv};
\addlegendentry{RC $160$}
\addplot [thick, Blues_6, mark=x, mark size=1.5, mark options={solid}] table [x=a, y=b,col sep=comma] {figures/Resnet1/resnet1.csv};
\addlegendentry{FCR $80$}
\addplot [thick, Reds_7, mark=*, mark size=1.5, mark options={solid}] table [x=a, y=e,col sep=comma] {figures/Resnet1/resnet1.csv};
\addlegendentry{RC $128$}
\end{axis}

\end{tikzpicture}

    }}
    \vspace{-25pt}
    \caption{Accuracy versus inference resolution $K_{eval}$ for varying training random crop (RC) resolutions $K_{train}$ and varying resized activations resolutions (FCR).}
    \label{fig:Resnet1}
\end{figure}
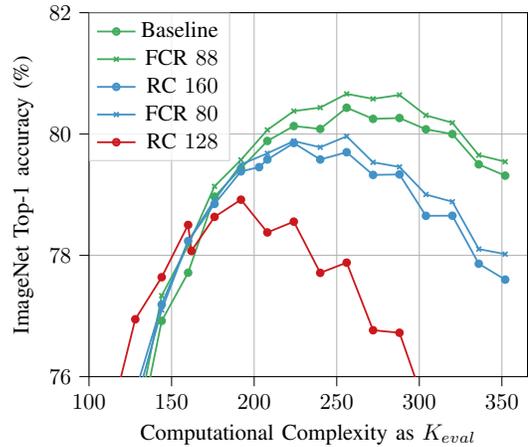

The results of those two experiments are shown on Figure~\ref{fig:Resnet1}. As observed in~\cite{touvronFixingTraintestResolution2022}, the maximum accuracy is obtained at a larger resolution than the one used while training the network due to a discrepancy between the training and evaluation data augmentations. We also observe that the second method outperforms the first method and the baseline for similar training costs. 

Additionally, we compare and apply our resolution scaling methods with model scaling. 

We evaluate the trained baseline on a range of $K$ and $K_{eval}$ and report the best compressed evaluation with a maximum of~0.75\% decrease in accuracy on ImageNet on Figure~\ref{fig:catchFigure}. 
Furthermore, we apply uniform model scaling to the ResNet-50 architecture (i.e. we multiply each convolution channels by a ratio, here 0.5 and 0.25) and we report the same metric on Figure~\ref{fig:catchFigure} with the FLOPs/memory search space of model scaling. Our results demonstrate that resolution scaling provides novel trade-offs in model compression that are not attainable through uniform model scaling, while maintaining competitive performance.

\subsubsection{Semantic Segmentation}\label{regseg}

For Semantic Segmentation, we select the RegSeg model  on the CityScapes dataset and we follow a similar experimental setup as for classification. 

We train the RegSeg architecture using a standard training routine, with transformations consisting of a random resize bounded by a lower value $LR$ and an upper value $HR$, followed by a random crop of size $RC$ with baseline values of $400$, $1600$, and $768$, respectively. 

During evaluation, the input is resized to a target resolution $K_{eval}$ and the output is interpolated to match the shape of the label in order to compute the mIoU on the validation set of CityScapes. Figure~\ref{SchemaRegSeg} depicts this evaluation pipeline with several resize resolution followed by two possible interpolations modes that are used to compare against the label.

\begin{figure}[h]
    \centering
\resizebox{0.5\textwidth}{!}{
    {
\begin{tikzpicture}

\definecolor{crimson2143940}{RGB}{214,39,40}
\definecolor{darkgray176}{RGB}{176,176,176}
\definecolor{darkorange25512714}{RGB}{255,127,14}
\definecolor{forestgreen4416044}{RGB}{44,160,44}
\definecolor{lightgray204}{RGB}{204,204,204}
\definecolor{mediumpurple148103189}{RGB}{148,103,189}
\definecolor{steelblue31119180}{RGB}{31,119,180}

\begin{axis}[
tick align=outside,
tick pos=left,
x grid style={darkgray176},
xlabel={Computational Complexity as $K_{eval}$},
xmajorgrids,
xmin=300, xmax=1606.4,
xtick style={color=black},
y grid style={darkgray176},
ylabel={Cityscapes mIoU},
ymajorgrids,
ymin=70, ymax=79,
ytick style={color=black}
]

\addplot [semithick, steelblue31119180, dotted, mark=*, mark size=1.5, mark options={solid}] table [x=a, y=b,col sep=comma]  {figures/RegSeg0/regseg0.csv};\label{pgfplots:regseg0}
\addplot [semithick, darkorange25512714, dotted, mark=square*, mark size=1.5, mark options={solid}] table [x=c, y=d,col sep=comma]  {figures/RegSeg0/regseg0.csv};\label{pgfplots:regseg1}
\addplot [semithick, forestgreen4416044, dotted, mark=triangle*, mark size=1.5, mark options={solid}] table [x=e, y=f,col sep=comma]  {figures/RegSeg0/regseg0.csv};\label{pgfplots:regseg2}
\addplot [semithick, crimson2143940, dotted, mark=diamond*, mark size=1.5, mark options={solid}] table [x=g, y=h,col sep=comma]  {figures/RegSeg0/regseg0.csv};\label{pgfplots:regseg3}
\addplot [semithick, mediumpurple148103189, dotted, mark=oplus*, mark size=1.5, mark options={solid}] table [x=i, y=j,col sep=comma]  {figures/RegSeg0/regseg0.csv};\label{pgfplots:regseg4}
\addplot [semithick, steelblue31119180, dotted, mark=pentagon*, mark size=1.5, mark options={solid}] table [x=k, y=l, col sep = comma] {figures/RegSeg0/regseg0.csv};\label{pgfplots:regseg5}
\addplot [semithick, darkorange25512714, dotted, mark=otimes*, mark size=1.5, mark options={solid}] table [x=m, y=n, col sep = comma] {figures/RegSeg0/regseg0.csv}; \label{pgfplots:regseg6}
\coordinate (legend) at (axis description cs:1.0,0.0);
\end{axis}

\matrix [
    draw,
    matrix of nodes,
    anchor=south east,
    fill=white,
    row sep=-2pt,
    column sep=-4pt,
    style={nodes={font=\small}}
] at (legend) {
    \pgfmatrixnextcell RC        \pgfmatrixnextcell LR               \pgfmatrixnextcell HR         \pgfmatrixnextcell FCR      \\ 
    \ref{pgfplots:regseg0} \pgfmatrixnextcell 1024 \pgfmatrixnextcell 400 \pgfmatrixnextcell 1600 \pgfmatrixnextcell \\
    \ref{pgfplots:regseg1} \pgfmatrixnextcell 768 \pgfmatrixnextcell 400 \pgfmatrixnextcell 1600 \pgfmatrixnextcell \\
    \ref{pgfplots:regseg2} \pgfmatrixnextcell 768 \pgfmatrixnextcell 300 \pgfmatrixnextcell 1200 \pgfmatrixnextcell \\
    \ref{pgfplots:regseg3} \pgfmatrixnextcell 512 \pgfmatrixnextcell 400 \pgfmatrixnextcell 1600 \pgfmatrixnextcell \\
    \ref{pgfplots:regseg4} \pgfmatrixnextcell 512 \pgfmatrixnextcell 200 \pgfmatrixnextcell 800 \pgfmatrixnextcell \\
    \ref{pgfplots:regseg5} \pgfmatrixnextcell 1024 \pgfmatrixnextcell 400 \pgfmatrixnextcell 1600 \pgfmatrixnextcell 256 \\
    \ref{pgfplots:regseg6} \pgfmatrixnextcell 1024 \pgfmatrixnextcell 400 \pgfmatrixnextcell 1600 \pgfmatrixnextcell 384 \\
};

\end{tikzpicture}

}
}
    \vspace{-15pt}
    \caption{RegSeg architecture resolution scaling training strategies on Cityscapes. Each label corresponds to Random Crop (RC), Low Resize Size (LR), High Resize (HR) and resize after the first convolutional layer (FCR) sizes. Orange corresponds to the baseline}
    \label{fig:RegSeg1}
\end{figure}
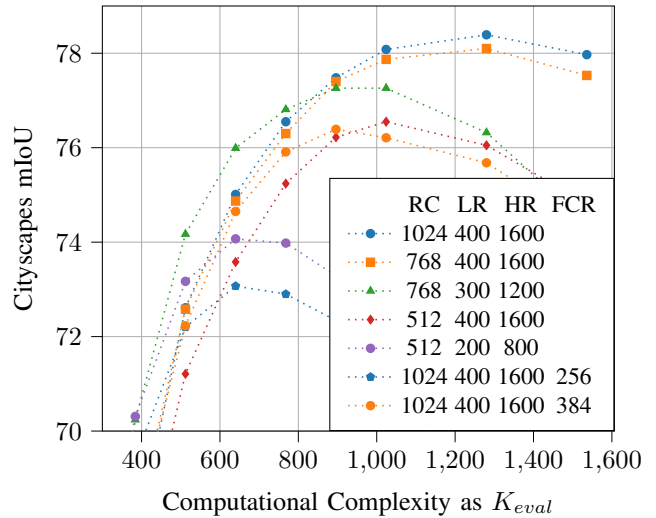

Similarly to the classification task, we evaluate the effectiveness our proposed methods.
We apply the same resolution scaling methods used for the ResNet-50 architecture to the RegSeg architecture to assess their effectiveness for semantic segmentation tasks. The first method is adapted because the RegSeg training routine does not use random resize crops. Instead, we evaluate the method by varying the random crop size $S_{crop}$ and the resize bounds $S_{low}$ and $S_{high}$. The second method is applied similarly to how it was applied to the ResNet-50 Architecture.
We evaluate both methods across a range of target resolutions $K$ and report the mIoU on the Cityscapes dataset for each configuration in Figure~\ref{fig:RegSeg1}. Our findings indicate that for the RegSeg architecture on Cityscapes, neither method significantly improves the mIoU. Finding the best compromise of the baseline with a grid search of the the evaluation crop size $K$ alone is sufficient to reduce the computational cost and the memory usage for activations.

We also compare and apply our resolution scaling methods with model scaling.

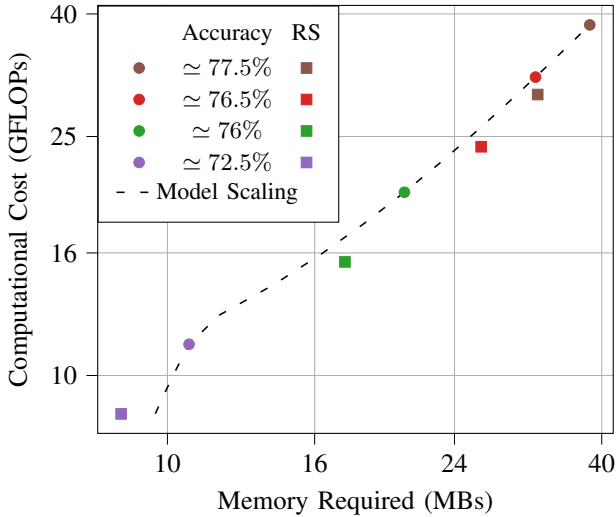
\begin{figure}
    \centering
    \begin{tikzpicture}

\definecolor{crimson2143940}{RGB}{214,39,40}
\definecolor{darkgray176}{RGB}{176,176,176}
\definecolor{darkorange25512714}{RGB}{255,127,14}
\definecolor{forestgreen4416044}{RGB}{44,160,44}
\definecolor{lightgray204}{RGB}{204,204,204}
\definecolor{mediumpurple148103189}{RGB}{148,103,189}
\definecolor{sienna1408675}{RGB}{140,86,75}
\definecolor{steelblue31119180}{RGB}{31,119,180}

\begin{axis}[
legend cell align={left},
legend style={
  fill opacity=0.8,
  draw opacity=1,
  text opacity=1,
  at={(0.03,0.97)},
  anchor=north west,
  draw=lightgray204
},
log basis x={10},
log basis y={10},
tick align=outside,
tick pos=left,
x grid style={darkgray176},
ylabel={Computational Cost (GFLOPs)},
xmajorgrids,
xmin=8004785.97371676, xmax=41496236.4999679,
xmode=log,
xtick style={color=black},
y grid style={darkgray176},
xlabel={Memory Required (MBs)},
ymajorgrids,
ymin=8010160306.07177, ymax=41326242334.72,
ymode=log,
ytick style={color=black},
ytick={10e9, 16e9 ,  25e9, 40e9},
yticklabels={10, 16, 25, 40},
xtick={10e6, 16e6, 25e6, 40e6},
xticklabels={ 10, 16, 24, 40},
scaled x ticks=false,
scaled y ticks=false,
]
\addplot [semithick, black, loosely dashed]
table{%
x  y
9622044 8638193104
9888820 9289938120
10253456 10167986354
10715452 11272337556
11799904 12602992130
14030568 14159949810
16359188 15943210756
18784816 17952774494
21308964 20188641780
23930652 22650812124
26650392 25339285782
29466788 28254062056
32382056 31395142054
35394768 34762525062
38505628 38356211432
};\label{pgfplots:regseg0_MS}
\addplot [draw=forestgreen4416044, fill=forestgreen4416044, mark=*, mark size=2, only marks]
table{%
x  y
21308964 20188641780
};\label{pgfplots:regseg0_0}
\addplot [draw=crimson2143940, fill=crimson2143940, mark=*, mark size=2, only marks]
table{%
x  y
32382056 31395142054
};\label{pgfplots:regseg0_1}
\addplot [draw=mediumpurple148103189, fill=mediumpurple148103189, mark=*,mark size=2,  only marks]
table{%
x  y
10715452 11272337556
};\label{pgfplots:regseg0_2}
\addplot [draw=sienna1408675, fill=sienna1408675, mark=*, mark size=2, only marks]
table{%
x  y
38505628 38356211432
};\label{pgfplots:regseg0_3}
\addplot [draw=forestgreen4416044, fill=forestgreen4416044, mark=square*, mark size=2, only marks]
table{%
x  y
17622564 15457024500
};\label{pgfplots:regseg0_4}
\addplot [draw=crimson2143940, fill=crimson2143940, mark=square*, mark size=2, only marks]
table{%
x  y
27221096 24037087654
};\label{pgfplots:regseg0_5}
\addplot [draw=mediumpurple148103189, fill=mediumpurple148103189, mark=square*,mark size=2,  only marks]
table{%
x  y
8626492 8630409876
};\label{pgfplots:regseg0_6}
\addplot [draw=sienna1408675, fill=sienna1408675, mark=square*, mark size=2, only marks]
table{%
x  y
32607388 29366709992
};\label{pgfplots:regseg0_7}
\coordinate (legend) at (axis description cs:0.0,1.0);
\end{axis}
\matrix [
    draw,
    matrix of nodes,
    anchor=north west,
    fill=white,
    row sep=-2pt,
    column sep=-9pt,
    style={nodes={font=\small}}
] at (legend) {
    \pgfmatrixnextcell Accuracy        \pgfmatrixnextcell RS             \\ 
    \ref{pgfplots:regseg0_3} \pgfmatrixnextcell $\simeq77.5\%$ \pgfmatrixnextcell \ref{pgfplots:regseg0_7}\\
    \ref{pgfplots:regseg0_1} \pgfmatrixnextcell $\simeq76.5\%$ \pgfmatrixnextcell \ref{pgfplots:regseg0_5}\\
    \ref{pgfplots:regseg0_0} \pgfmatrixnextcell $\simeq76\%$ \pgfmatrixnextcell \ref{pgfplots:regseg0_4}\\
    \ref{pgfplots:regseg0_2} \pgfmatrixnextcell $\simeq72.5\%$ \pgfmatrixnextcell \ref{pgfplots:regseg0_6} \\
    \ref{pgfplots:regseg0_MS} \pgfmatrixnextcell Model Scaling   \\
};

\end{tikzpicture}
    \vspace{5mm}
    \caption{Resolution scaling applied to a RegSeg in addition to
Model Scaling. Each color corresponds to a scaled model that has
been compressed with resolution scaling as much as possible with a
0.75 \% drop in mIoU on the Cityscapes dataset.}
    \label{fig:regseg0}
\end{figure}

As RegSeg architectures uses grouped convolutions that severely limits model scaling with a uniform reduction of channels, we select the group width $gw$ as the scaling factor and keep the number of groups per filters constant, improving the search space attainable by model scaling. We report the same metric as on the baseline in Figure~\ref{fig:regseg0}.

Our results demonstrate that resolution scaling enables novel trade-offs that are not achievable with model scaling, maintaining competitive performance. Specifically, resolution scaling reduces the required memory for activations and FLOPs by~24\%, with only a~0.6\% decrease in mIoU on the Cityscapes dataset when compared to the baseline. The RegSeg architecture, using grouped and dilated convolutions, presents limited opportunities for model scaling, which restricts its effectiveness within the search space. Therefore, resolution scaling proves to be more versatile, extends the limited search space available for the RegSeg architecture, allowing for a broader range of networks with varying computational and memory costs, making it suitable for a wider array of applications.

\subsection{Vision Transformers}
We investigate the effects of sequence scaling and model scaling on Vision Transformers (ViTs). The experiments are based on the ViT-Small architecture, chosen for its relatively simple and fast training process. All networks were trained using the torchvision library.

\begin{figure}[hbp]
    \centering
    \resizebox{0.4\textwidth}{!}
{
    {        
    \begin{tikzpicture}
    
    \definecolor{darkgray176}{RGB}{176,176,176}
    \definecolor{green}{RGB}{0,128,0}
    \definecolor{lightgray204}{RGB}{204,204,204}
    \definecolor{forestgreen4416044}{RGB}{44,160,44}
    
    \begin{axis}[
    tick align=outside,
    tick pos=left,
    x grid style={darkgray176},
    xlabel={Computational Cost (GFLOPs)},
    xmajorgrids,
    xmin=4099729796.4, xmax=17700172467.6,
    xtick style={color=black},
    y grid style={darkgray176},
    ylabel={ImageNet Top-1(\%)},
    ymajorgrids,
    ymin=70.3647, ymax=76.4653,
    ytick style={color=black},
    xtick={0.0,6000000000, 8000000000, 10000000000, 12000000000, 14000000000, 16000000000, 18000000000},
    xticklabels={0.0,6, 8, 10, 12, 14, 16, 18},
    scaled x ticks=false,
    ]
    
    \addplot [draw=Set1_3, fill=Set1_3, mark=*, only marks] table [x=c, y=d,col sep=comma]   {figures/VIT1/vit1.csv};\label{plot:seq64}
    \addplot [draw=Set1_4, fill=Set1_4, mark=*, only marks] table [x=e, y=f,col sep=comma]   {figures/VIT1/vit1.csv};\label{plot:seq121}
    \addplot [draw=Set1_5, fill=Set1_5, mark=*, only marks] table [x=g, y=h,col sep=comma]   {figures/VIT1/vit1.csv};\label{plot:seq144}
    \addplot [draw=Accent_5, fill=Accent_5, mark=*, only marks] table [x=i, y=j,col sep=comma]   {figures/VIT1/vit1.csv};\label{plot:seq169}
    \addplot [draw=Set1_7, fill=Set1_7, mark=*, only marks] table [x=k, y=l,col sep=comma]   {figures/VIT1/vit1.csv};\label{plot:seq196}
    \addplot [draw=Set1_8, fill=Set1_8, mark=*, only marks] table [x=m, y=n,col sep=comma]   {figures/VIT1/vit1.csv};\label{plot:seq225}

    \addplot [semithick, black, dashed, forget plot] table [x=a, y=b,col sep=comma]  {figures/VIT1/vit1.csv};
    \coordinate (legend) at (axis description cs:0.0,1.0);
    \end{axis}

\matrix [
    draw,
    matrix of nodes,
    anchor=north west,
    fill=white,
    row sep=-2pt,
    column sep=-4pt,
] at (legend) {
    \pgfmatrixnextcell Sequence              \\
     \pgfmatrixnextcell Length              \\
    \ref{plot:seq64} \pgfmatrixnextcell 64    \\
    \ref{plot:seq121} \pgfmatrixnextcell 121   \\
    \ref{plot:seq144} \pgfmatrixnextcell 144   \\
    \ref{plot:seq169} \pgfmatrixnextcell 169   \\
    \ref{plot:seq196} \pgfmatrixnextcell 196   \\
    \ref{plot:seq225} \pgfmatrixnextcell 225   \\
};
    \end{tikzpicture}

    }
}

    \vspace{-5pt}
    \caption{Impact of the input resolution with a fixed patch size (16×16) for a ViT-S}
    \label{fig:vit1}
\end{figure}
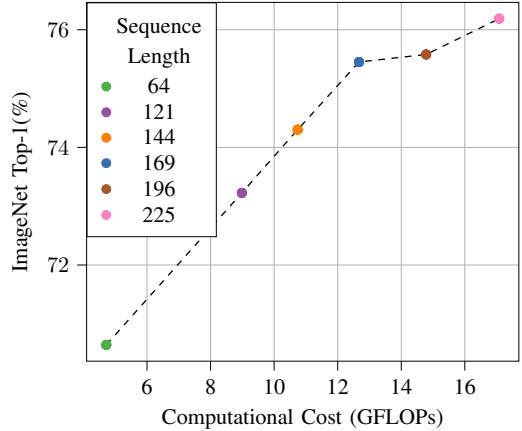

We first explore the impact of resolution scaling by training and evaluating ViTs across a range of resolutions \{8, 11, 12, 13, 14, 15\}, keeping the patch size fixed at 16x16. This allows us to assess the effect of varying resolution, and by extension image resolution, on both the accuracy and computational cost of the model during training and evaluation. The results are shown on Figure~\ref{fig:vit1} and show that reducing the resolution has a limited impact on accuracy. Specifically, we observe only a~1\% drop in accuracy while achieving a~28\% reduction in FLOPs when comparing the baseline with a resolution of 14 tokens per column or line.



However, Increasing the resolution at a fixed patch size implies training and evaluating the ViT with a higher resolution image that can lead to a mismatch between the optimal image resolution and the optimal architecture resolution (i.e. tokens per column/line). In order to evaluate the relationship between the patch size and the architecture resolution, we train and evaluate ViT-S at two different resolutions \{9, 11\} and multiple patch size \{8x8, 12x12, 16x16, 24x24, 32x32\}, the results are shown on Table~\ref{tab:vit0}. The best results for an architecture resolution of 9 and 11 are respectively obtained with patch size of 16x16 and 12x12, we can observe that the image resolution obtained with these resolution are similar with respectively 128x128 and 132x132 showing the importance of evaluating the best image size for the task and dataset selected. 

\begin{table*}
\begin{minipage}{\columnwidth}
    \centering
    \begin{tabular}{|c|c|c|c|c|}
    \multicolumn{5}{c}{\textit{Sequence Length 9}}  \\ \hline
    \multicolumn{2}{|c|}{\multirow{2}{*}{\textbf{Method}}}     & \textbf{Memory}  & \textbf{FLOPs} & \multirow{2}{*}{\textbf{Acc}} \\ 
    \multicolumn{2}{|c|}{}      &  \textbf{Required} (MBs) & (GFLOPs) &   \\ \hline
    \multicolumn{2}{|c|}{\multirow{2}{*}{\shortstack{Hidden Size \\ Scaling }}} &  \multirow{2}{*}{25.9}  & \multirow{2}{*}{4.5} & \multirow{2}{*}{47.54}           \\
    \multicolumn{2}{|c|}{} & &  & \\ \hline
    \multicolumn{2}{|c|}{MLP Scaling}         & 66.6 & 4.7  &  \underline{68.89}         \\ \hline
    \multicolumn{2}{|c|}{Depth Scaling}        & 65.3 & 5.0  &  62.14         \\ \hline
    \multicolumn{2}{|c|}{Hybrid Scaling}        & 39.4 & 4.8  &  67.64        \\ \hline
    \multirow{5}{*}{\shortstack{Resolution Scaling}} 
                              & 8  &   26.1 & 4.7  &  69.16           \\ \cline{2-5}
                              & 12  &   32.0 & 4.7  &  69.77          \\ \cline{2-5}
                              & 16  &   40.3 & 4.7  &  \textbf{70.64}           \\ \cline{2-5}
                              & 24  &   63.9 & 4.8  &  68.55          \\ \cline{2-5}
                              & 32  &   97.0 & 4.8  &  69.31         \\ \hline
    \end{tabular}
\end{minipage}\hfill 
\begin{minipage}{\columnwidth}
    \centering
    \begin{tabular}{|c|c|c|c|c|}
    \multicolumn{5}{c}{\textit{Sequence Length 11}}  \\ \hline
    \multicolumn{2}{|c|}{\multirow{2}{*}{\textbf{Method}}}     & \textbf{Memory}  & \textbf{FLOPs} & \multirow{2}{*}{\textbf{Acc}} \\ 
    \multicolumn{2}{|c|}{}      &  \textbf{Required} (MBs) & (GFLOPs) &   \\ \hline
    \multicolumn{2}{|c|}{\multirow{2}{*}{\shortstack{Hidden Size \\ Scaling }}} & \multirow{2}{*}{50.7}  & \multirow{2}{*}{9.1} & \multirow{2}{*}{71.02}           \\
    \multicolumn{2}{|c|}{} & &  & \\ \hline
    \multicolumn{2}{|c|}{MLP Scaling}         & 72.1 & 9.0  &  \textbf{73.68}          \\ \hline
    \multicolumn{2}{|c|}{Depth Scaling}       & 70.6 & 8.7  &  70.92         \\ \hline
    \multirow{5}{*}{\shortstack{Resolution Scaling}} 
                              & 8  &   30.4 & 8.9  &  73.21          \\ \cline{2-5}
                              & 12  &   41.6 & 9.0  &  \underline{73.48}          \\ \cline{2-5}
                              & 16  &   57.2 & 9.0  &  73.23                            \\ \cline{2-5}
                              & 24  &   101.8 & 9.1  &  72.89       \\ \cline{2-5}
                              & 32  &   164.4 & 9.2  &  72.98         \\ \hline
    \end{tabular}
  \end{minipage}
  \caption{Comparison of scaling methods for ViTs at two computational cost. The top-1 accuracy on ImageNet is reported for each method. For resolution scaling, multiple patch size are evaluated with a resolution of 9 and 11 tokens per column/line. For Model Scaling, individual scalings are considered for the MLP size, the hidden size and the depth size. An hybrid scaling corresponding to the scaling of a ViT-B to a ViT-S is also evaluated}
  \label{tab:vit0}
\end{table*}

Additionally, we compare resolution scaling and model scaling for ViTs using two resolutions, 9 and 11 tokens per column/line with multiple patch sizes \{8x8, 12x12, 16x16, 24x24, 32x32\} and for each ViT architecture parameter, $D_{MLP}$, $D$, $n$ we scale a ViT in order to match the number of FLOPs of a base ViT at each resolution. The result are reported in Table~\ref{tab:vit0} alongside the number of FLOPS and the memory required for each configuration. These results show that resolution scaling is more effective than model scaling in limited computational cost scenario and is effective compared to resolution scaling is the MLP scaling that is superior at the higher resolution and is inferior compared to each patch size at a lower resolution.

\subsection{Quantization}\label{quantif}

In this section, we investigate whether the use of quantization is complementary or not to resolution scaling and to model scaling.
We apply 8-bits integer static quantization to each trained ResNet and RegSeg models using a calibration dataset that is based on the training dataset and the specific resolutions that were used in the training phase of each model. We evaluate each models using the same methodology used in \ref{resnet} and \ref{regseg}. We report the results for a baseline model in Table~\ref{quantresnet} and \ref{quantregseg}.

\begin{table*}
\begin{minipage}{\columnwidth}
    \centering
    \begin{tabular}{|c|c|c|c|}
    \multicolumn{4}{c}{Resnet} \\ \hline
    \multirow{2}{*}{\textbf{Resolution}} & \multirow{2}{*}{\textbf{Baseline}} & \textbf{Quantized} & \multirow{2}{*}{\textbf{Difference}} \\
     & & \textbf{Model} & \\ \hline
    120  &  72.89  &  72.08 & 0.81 \\ \hline
    152  &  77.11  &  76.52 & 0.59 \\ \hline
    184  &  79.22  &  78.54 &  0.68 \\ \hline
    216  &  80.1  &  79.65  & 0.45 \\ \hline
    248  &  80.12  &  79.75 & 0.37 \\ \hline
    280  &  80.11  &  79.6  & 0.51 \\ \hline
    312  &  79.86  &  79.0 & 0.86 \\ \hline
    344  &  79.19  &  78.66 & 0.53 \\ \hline
    \end{tabular}
    \caption{Performance comparison of the baseline ResNet model and its quantized version across different input resolutions on the ImageNet dataset. The "Difference" column indicates the absolute difference in accuracy}
    \label{quantresnet}
\end{minipage}\hfill 
\begin{minipage}{\columnwidth}
    \centering
    \begin{tabular}{|c|c|c|c|}
    \multicolumn{3}{c}{Regseg} \\ \hline
    \multirow{2}{*}{\textbf{Resolution}} & \multirow{2}{*}{\textbf{Baseline}} & \textbf{Quantized} & \multirow{2}{*}{\textbf{Difference}} \\
     & & \textbf{Model} & \\ \hline
    128  &  31.23  &  30.35  & 0.88 \\ \hline
    384  &  67.61  &  67.01  & 0.6  \\ \hline
    640  &  74.89  &  74.3   & 0.59 \\ \hline
    896  &  77.53  &  76.89  & 0.64 \\ \hline
    1024  & 77.99  &  77.44  & 0.55 \\ \hline
    1280  & 78.37  &  77.64  & 0.73 \\ \hline
   \end{tabular}
    \caption{Performance comparison of the baseline RegSeg model and its quantized version across different input resolutions on the CityScapes dataset. The "Difference" column indicates the absolute difference in mIoU}
    \label{quantregseg}
  \end{minipage}

\end{table*}
These results show that quantization is complementary to the use of resolution scaling as it affects similarly each model in the same way at every resolution.
\section{Conclusion} 
In this study, we have introduced novel mechanisms for modifying input resolution, both before and after the embedding process in ResNets and for sequence length reduction in Vision Transformers. Through systematic investigation, we demonstrated the potential of input resolution reduction as an effective standalone compression technique and as a complementary method to existing approaches such as model scaling and quantization. Our extensive experiments highlighted the consistent ability of this approach to achieve a good trade-off between computational cost and performance across diverse tasks and architectures.

These findings emphasize the critical role of input resolution as a practical and impactful dimension in the model compression landscape for computer vision. By integrating resolution modification into compression strategies, we open pathways for further research and optimization in designing efficient and high-performing models.


\section*{Acknowledgments}
This research was funded, in whole or in part, by the French National Research Agency (ANR) under the project ANR-22-CE25-0006 and was performed using AI resources from
GENCI-IDRIS.

\bibliographystyle{IEEEtran}
\bibliography{ijcnn}

\end{document}